\documentclass[runningheads]{llncs}
\usepackage[T1]{fontenc}
\usepackage{amsmath}
\usepackage{graphicx}
\usepackage{color}
\usepackage{tabularx}
\usepackage{booktabs}
\usepackage{subcaption}
\newcolumntype{C}{>{\centering\arraybackslash}X}
\usepackage{hyperref}

\begin{document}

\title{Root Causing Prediction Anomalies Using Explainable AI}
\titlerunning{Root Causing Prediction Anomalies Using XAI}
\author{Ramanathan Vishnampet\inst{1} \and Rajesh Shenoy\inst{1} \and Jianhui Chen\inst{1} \and Anuj Gupta\inst{1}}
\authorrunning{R. Vishnampet et al.}
\institute{Meta, 1 Hacker Way, Menlo Park, CA 94065, USA\\
\email{\{rvishna,rshenoy,jchen2020,ag1\}@meta.com}}

\maketitle

\begin{abstract}
This paper presents a novel application of explainable AI (XAI) for root-causing performance degradation in machine learning models that learn continuously from user engagement data. In such systems a single feature corruption can cause cascading feature, label and concept drifts. We have successfully applied this technique to improve the reliability of models used in personalized advertising. Performance degradation in such systems manifest as prediction anomalies in the models. These models are typically trained continuously using features that are produced by hundreds of real time data processing pipelines or derived from other upstream models. A failure in any of these pipelines or an instability in any of the upstream models can cause feature corruption, causing the model’s predicted output to deviate from the actual output and the training data to become corrupted. The causal relationship between the features and the predicted output is complex, and root-causing is challenging due to the scale and dynamism of the system. We demonstrate how temporal shifts in the global feature importance distribution can effectively isolate the cause of a prediction anomaly, with better recall than model-to-feature correlation methods. The technique appears to be effective even when approximating the local feature importance using a simple perturbation-based method, and aggregating over a few thousand examples. We have found this technique to be a model-agnostic, cheap and effective way to monitor complex data pipelines in production and have deployed a system for continuously analyzing the global feature importance distribution of continuously trained models.

\end{abstract}

\section{Introduction}
\label{sec:introduction}

Performance degradation of machine learning models after being deployed to use is a well known problem faced by machine learning practitioners~\cite{Sculley2015,Kenthapadi2022}. The causes of degradation have been broadly classified as concept drift~\cite{Zhang2023,Joao2014} or data drift~\cite{Kenthapadi2022}. Concept drift occurs when the relationship between input and target variable changes over time. Concept drift is also referred to as model staleness. Data drift occurs if the input or target distribution changes over time. Input changes are called feature drifts and target changes are called label drifts. Supervised and unsupervised statistical approaches have been proposed to detect both concept and data drift~\cite{Budhathoki2021}.  With the increasing interest in model explainability in the last few years, a new area of drift detection using explainability approaches~\cite{He2023,Taly2021,Nigenda2022,Decker2023} has emerged. Compared to statistical approaches, explainability based approaches also hold the promise to help the practitioner identify and mitigate the causes of degradation.

We describe the unique challenges of monitoring machine learning based recommendation systems that learn continuously from user engagement~\cite{He2014}. Such systems have an initial training phase after which they are deployed in production. Once deployed in production, these systems get new engagement labels and are trained regularly based on these labels. One would hope such online training would seemingly avoid the problems of concept drift. However, being in the trenches of monitoring such systems, we have identified that these systems are susceptible to data (both label and feature) and concept drifts.

To illustrate this problem, consider a model which is deployed in production that predicts the likelihood of engagement. Such systems use a myriad of signals (called input features) to make the prediction. Once the prediction is made, the system chooses to display the result which has the highest predicted value. Assume that due to a single feature corruption, the model now predicts a larger value. With such a value the result would have a higher chance of being displayed but have a lower chance of engagement. This causes label drift where the new labels have a different distribution than before the feature corruption. The displayed item along with its uncorrupted features now also enters into the training pipeline for training the next iteration of the model. Since the item would not have been displayed in the absence of the feature corruption, this also creates a concept drift in the model which can quickly cascade and cause severe degradations in user experience or business outcome.

Thus in such systems, detecting feature corruption and mitigating them as soon as possible becomes paramount. The problem is further complicated since in a typical production deployment there could be hundreds of such models each with thousands of features with some model outputs being used as inputs to other models. We use techniques from the explainability area to tackle this problem.

\section{Related Work}
\label{sec:related-work}

Explainability based performance monitoring has been proposed in~\cite{Koebler2023}. In their approach, optimal transport is used to detect distribution shifts and the Shapley method is used to identify the important features. In contrast, for our use case, we can directly determine the distribution shift from actual click data and our problem is to identify which features caused the problem, if any. They demonstrated their approach on the MNIST dataset with simulated corruptions. In another work, feature importance was applied to tabular data~\cite{Mougan2023}. Explainability and monitoring of models are considered as separate concerns in~\cite{Klaise2020} while we use explainability techniques directly for model monitoring.

\section{Problem Statement}
\label{sec:problem-statement}

For large scale user engagement models deployed in production, we can calculate the actual engagement events observed over a window of time. The discrepancy from actual to predicted events is used as a signal that a degradation has occurred. The calibration of the prediction is the ratio of the number of expected events to the number of actually observed events. Significant deviations from a calibration value of $1$ are flagged as anomalies and initiate a triaging process. Several techniques are used to triage the cause of the degradation, which could range from user behavioral changes, infrastructure issues, model issues or feature corruption. We want to come up with an approach that helps the triaging to identify the features which could have caused the anomaly.

\section{Methodology}
\label{sec:methodology}

This section describes our methodology to triage model performance degradation to feature corruption. Briefly, the steps involved are:

\begin{enumerate}
\item Estimate local feature importances (LFI) on the \emph{control} and \emph{anomaly} data, i.e., before and during the event, respectively.
\item Aggregate LFIs to obtain global feature importances (GFIs).
\item Rank features based on the shift in GFI between control and anomaly data. The ranked list is used to prioritize the triaging process.
\item Triage the top features from the ranked list further to root cause the anomaly to a broken pipeline, model drift in an upstream model, etc., and take appropriate short-term and long-term mitigation actions.
\end{enumerate}

Step 1 involves running inference on several thousands to millions of examples using the model, and is the most computationally expensive. Steps 1, 2 and 3 can be fully automated and we will later describe a system that performs these steps continuously instead of retroactively after a suspected performance degradation. In principle, step 4 can be automated to establish a self-healing system -- for example, through an automated runbook that applies mitigation actions by selecting features based on their GFI shift through a heuristic (e.g., thresholding) or learning algorithm. In practice, experts are involved to supervise step 4 to catch false positives.

\subsection{Local Feature Importance Estimation}
\label{sec:lfi-estimation}

\begin{figure*}[t!]
    \centering
    \begin{subfigure}[t]{0.55\linewidth}
        \centering
        \includegraphics[width=\linewidth]{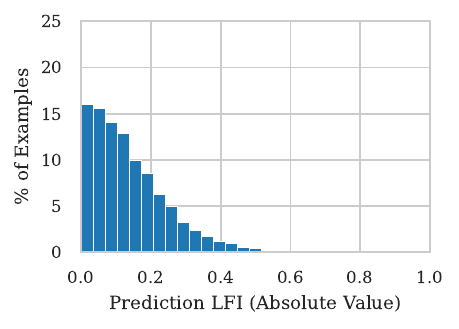}
        \caption{Based on relative prediction change from baseline~\eqref{eq:prediction-lfi}}
    \end{subfigure} \\
    \begin{subfigure}[t]{0.55\linewidth}
        \centering
        \includegraphics[width=\linewidth]{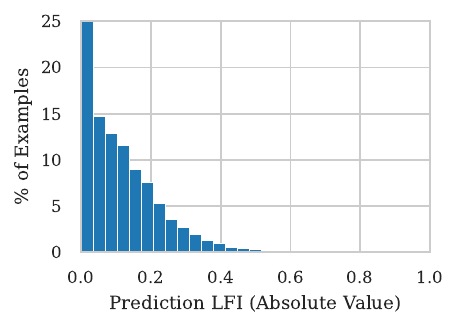}
        \caption{Based on pseudo loss change from baseline~\eqref{eq:pseudo-loss-lfi}}
    \end{subfigure}
    \caption{Distribution of absolute values of LFIs for a particular feature using the definitions in~\eqref{eq:prediction-lfi} and~\eqref{eq:pseudo-loss-lfi}}
    \label{fig:lfi-histograms}
\end{figure*}

Consider a model $\mathcal{M}$ that takes an input example $X_i$ and predicts the probability of an event $p_i=\mathcal{M}\left(X_i\right)$. $X_i$ is joined with the label $y_i\in\{0, 1\}$ (whether an actual event was observed) by an ETL (extract, transform and load) process and stored as a training example in a data warehouse -- a process that takes several minutes to a few hours. To facilitate faster root-causing and debugging data integrity problems, we log a subset of input examples without labels. LFIs are computed on the unlabeled data sampled from the time of the event (anomaly) and a time before the event (control). We load a checkpoint of the model that hasn't seen either the control or the anomaly data during training in an isolated hardware environment for inference. The hardware isolation avoids confounding effects from infrastructure bugs that may be the underlying cause of the anomaly. The same checkpoint is used for both control and anomaly data to avoid confounding effects from checkpoint corruption. We use the Feature Ablation algorithm available in the open-source Captum library~\cite{Kokhilyan2020} with a choice of baseline described in section~\ref{sec:choice-of-baseline}, and compute the LFI as
\begin{equation}
\label{eq:pseudo-loss-lfi}
    w_{ij} = \begin{cases}
\log\left(\frac{\tilde{p}_{ij}}{p_i}\right), & \text{if } y_i^\prime= 1\\[0.5em]
\log\left(\frac{1 - \tilde{p}_{ij}}{1 - p_i}\right),         & \text{otherwise}
\end{cases}
\end{equation}
\noindent where $\tilde{p}_{ij}$ is the prediction when the feature $x_{ij}$ in example $X_i$ is replaced by its baseline value, and $y_i^\prime$ is a ``pseudo label'' computed by applying a fixed threshold to the prediction $p_i$. This definition is a slight departure from the attribution score computed by the Feature Ablation algorithm in Captum
\begin{equation}
\label{eq:prediction-lfi}
w_{ij} = 1 - \tilde{p}_{ij} / p_i
\end{equation}
Our definition of the LFI is motivated by the binary cross entropy loss function typically used for training such models. Figure~\ref{fig:lfi-histograms} compares the distribution of absolute values of LFIs based on the two definitions for a particular feature. Binary cross entropy loss does not heavily penalize the actual value of the prediction when it is close to the true label, but has a sharp gradient when the prediction is away from the true label. Hence, the LFI distribution based on pseudo loss is more skewed towards smaller values.

\subsection{Choice of Baseline}
\label{sec:choice-of-baseline}

It is well established that attribution methods are sensitive to the choice of baseline~\cite{Haug2021}. Our primary objective here, however, is not the accuracy of the LFIs, but rather the precision and recall of identifying features causing a prediction anomaly. Our choice of baseline is motivated by the following principles:

\begin{enumerate}
    \item The same baseline must be chosen for both the control and anomaly data. This is because the GFIs for each feature corresponding to the control and anomaly data must be comparable. The GFIs are obtained by independently aggregating the LFIs, which must therefore be computed relative to the same baseline.
    \item The process of obtaining the baseline value for a feature should be simple and computationally cheap compared to the attribution method itself.
    \item The baseline value for a feature should not depend on the local distribution of data at any given time. This is because the data distribution shows gradual and seasonal shifts that are expected but non-homogeneous across features, and choosing a data-dependent baseline can incorrectly pinpoint features due to shifts in their baseline value relative to others.
\end{enumerate}

Based on these considerations, we choose a static and deterministic baseline value for each feature. For numerical features, the baseline is a fixed feature-dependent value. For categorical features, the baseline is a special value that represents an ``unknown'' category. For embedding features, the baseline is the null vector in the embedding vector space. These choices are consistent with the ``default'' value of a feature used in place of missing data during training. Hence, the baseline values are not outside the training data distribution. Further, this choice of baseline guarantees that the LFI of features with default values is identically zero.

Seasonality in feature value distribution also affects the choice of the control data. We experimented with two ways of selecting the control -- from the previous hour relative to the anomaly data, and from the same hour on the previous day relative to the anomaly data. We found that the latter minimizes seasonality-induced noise in the LFI distribution for features with high seasonality.

\subsection{Global Aggregation}
\label{sec:global-aggregation}

To explain the global behavior of the model, we aggregate the LFIs to obtain GFIs. We express the LFIs as an $N\times M$ matrix $\mathbf{W}=\left[W_{ij}\right]$, where $N$ is the number of examples and $M$ is the number of features used by the model. We define the coverage of a feature to be the fraction of non-zero elements along the corresponding column of $\mathbf{W}$
\begin{equation}
    c_{ij}=\left|\left\{i:w_{ij}\neq 0\right\}\right|/N
\end{equation}

There are three considerations for aggregation:
\begin{enumerate}
    \item How should we choose the set of examples?
    \item How should we handle the sign of the LFI scores  when aggregating?
    \item Should features with higher coverage be assigned higher global importance and vice-versa?
\end{enumerate}

For the set of examples to aggregate over, we sample uniformly from two distributions of data with equal weight. The first is the set of examples on which the model makes a prediction. This includes all items that have been considered for ranking, including those that are not eventually displayed. The second is the set of examples where the item was displayed. Overall, this results in a higher weight for the set of examples that resulted in displayed items. The rationale for equally weighting these two distributions is (a) to capture the global importance of features that affect the prediction during inference, and (b) to understand which features have a larger impact on the data used for continuous online training of the model. The latter can cause feature and label drift which cascades into concept drift by affecting the training of subsequent checkpoints of the model.

LFIs include both positive and negative values. Averaging the LFIs with the sign may inaccurately diminish a feature’s importance. While it is sensible to average the absolute value of the LFIs, this is not an obvious choice. Theoretically, it is possible that directional shifts such as a change in the feature value distribution that causes its LFIs to flip sign are missed when identifying corrupted features. In practice, such a scenario is uncommon. Therefore, we make an empirical and practical choice of averaging the absolute value of LFIs.

Lastly, we consider features with low coverage. Their LFIs may be identically zero either because of missing values (which default to the baseline values), or because the prediction does not change when they are replaced by their baseline values. In either case, their overall impact to the global model behavior is small, which should be reflected in how their LFIs are aggregated.

Based on these considerations, we define the global importance of a feature as
\begin{equation}
    I_j = c_j \textrm{med}\left(\left\{\left|w_{ij}\right|:w_{ij}\neq 0\right\}\right)
\end{equation}
\noindent where $\textrm{med}$ denotes the median. We use the median instead of an average to handle outliers in the LFI distribution. We also exclude identically zero LFI values when computing the median to avoid identically zero global importance values for features with coverage lower than 50\%, instead choosing to weight the median of non-zero local importance values with the coverage.

How should we choose the sample size $N$ for aggregation? The objective is to choose a sufficiently large value such that the sample distribution of LFIs approximates the population distribution well. We determined this by computing the standard errors of the GFI estimates as defined above using bootstrap resampling on $1000$ samples each for various values of $N$. We picked a value that reduced the standard error across all features to below an acceptable tolerance threshold. We repeated this analysis for multiple datasets and models and identified a sufficiently large value of $N$ which we then fixed for all calculations. We found the optimal $N$ to be model-dependent, but conservatively use the largest value of $N=10,000$ across all models.

Once we have determined GFIs using the control and anomaly data, features are ranked based on the absolute value of the shift in GFIs between the two. The top $K$ features are surfaced to an expert who triages them based on their lineage and a timeline view of related events. Two other secondary metrics that are surfaced to the expert include the change in coverage of the features, and the rank shift based on GFI between the control and anomaly data. 

\subsection{Model-Feature Correlation}
\label{sec:mfc}

Model-feature correlation (MFC) is an alternative statistical approach for ranking features based on the shift in their Pearson correlation coefficient with the prediction between the control and anomaly data. The MFC score of a feature is defined as
\begin{equation}
    I_j = \frac{\left|\sum_{i=1}^N\left(x_{ij}-\bar{x}_j\right)\sum_{i=1}^N\left(p_{i}-\bar{p}\right)\right|}{\sqrt{\sum_{i=1}^N\left(x_{ij}-\bar{x}_j\right)^2\sum_{i=1}^N\left(p_i-\bar{p}\right)^2}}
\end{equation}
\noindent where $\bar{x}_j$ is the average feature value and $\bar{p}$ is the average prediction. For non-numerical features, we use a proxy value, e.g., the number of IDs for a sparse ID list feature, and the sum of weights of IDs for a weighted sparse ID list feature. Unlike the XAI approach, MFC does not require running inference on the model, which makes it attractive as it is cheaper and operates solely on already logged data. For example, with an aggregation sample size $N=10,000$ and a model that uses $\mathcal{O}\left(1000\right)$ features, computing GFIs for the control and anomaly data requires approximately $20$ million forward passes over the model. The advantage of the XAI approach is that it is based on the causal relationship between the model and features, and is therefore, more likely to point to the actual root cause with high precision and recall. We confirm this hypothesis with our results in section~\ref{sec:results}.

\section{Results}
\label{sec:results}

We modeled three broad categories of feature corruptions based on our experience and curated 11 cases to test the effectiveness of our XAI approach with MFC. The first 10 cases simulate different corruption types on the same data, which serves as the control. The last case is based on a real feature corruption that we triaged successfully.
\begin{enumerate}
    \item \textbf{Value distribution change}: For a numerical feature, value distribution change can occur due to a bug in a pipeline or concept drift in an upstream model that computes the feature. Cases 1 and 2 simulate a linear and non-linear change in a numerical feature, and case 3 simulates a categorical feature which is set to a constant value that is different from the baseline value of ``unknown'' category.
    \item \textbf{Coverage drop}: Cases 4 and 6 simulate a total coverage drop where no data is being populated for a subset of features. Cases 5 and 7 simulate a partial coverage drop that can occur when multiple pipelines are involved in populating a feature and some of them are broken. In cases 4 and 5, the features with simulated coverage drop are randomly selected, whereas in cases 6 and 7, they are selected based on their prior importance measured using their GFIs on the control data. Cases 8, 9 and 10 are also coverage drop corruptions, but for sparse features, which are lists of IDs that are transformed into real-valued dense features through embedding table lookups.
    \item \textbf{Encoded embedding corruption}: Some sparse features are encoded embedding representations~\cite{Zhang2023_1} that are transformed to real-valued dense representations by decoding the features at inference time. They are more susceptible to value distribution change from corruptions in their encoded representation. In case 11, we simulate corruption on 2 sparse-encoded embedding features.
\end{enumerate}

\begin{table}
\caption{Comparison of root causing effectiveness between XAI and MFC approaches on various feature corruption types}
\label{tbl:gfi-mfc-precision-recall}
\renewcommand{\arraystretch}{1.5}
\setlength{\tabcolsep}{3pt}
\begin{tabularx}{\linewidth}{
>{\hsize=0.1\hsize\raggedleft\arraybackslash}X
>{\hsize=1.5\hsize}X
>{\hsize=0.5\hsize}C
>{\hsize=0.5\hsize}C
>{\hsize=0.5\hsize}C
>{\hsize=0.5\hsize}C
}
\toprule
\# & Corruption Type & Features Corrupted & Average Prediction Change from Control & Corrupted Features in Top $K$ (GFI Shift) & Corrupted Features in Top $K$ (MFC Shift) \\
\midrule
 $1$ & $x \to 2x$ (linear change) & 
 $1$ & $-0.78\%$ & $1$ & $1$ \\
 
 $2$ & $x \to x^3$ (non-linear change) & 
 $1$ & $-7.28\%$ & $1$ & $0$ \\
 
 $3$ & Set categorical feature to constant value different from baseline value & 
 $1$ & $ 9.27\%$ & $1$ & $0$ \\
 
 $4$ & Replace $3$ randomly selected features with baseline value for $100\%$ of examples & 
 $3$ & $-5.33\%$ & $2$ & $1$ \\
 
 $5$ & Replace $3$ randomly selected features with baseline value for $50\%$ of examples & 
 $3$ & $-5.23\%$ & $1$ & $3$ \\
 
 $6$ & Replace $2$ important and $2$ unimportant features with baseline value for $100\%$ of examples & 
 $4$ & $-8.01\%$ & $2$ & $0$ \\
 
 $7$ & Replace $2$ important and $2$ unimportant features with baseline value for $50\%$ of examples & 
 $4$ & $-6.42\%$ & $2$ & $1$ \\
 
 $8$ & Randomly remove $50\%$ of IDs from sparse ID list feature & 
 $1$ & $ 0.12\%$ & $1$ & $0$ \\
 
 $9$ & Remove most recent $50\%$ of IDs from sparse ID list feature & 
 $1$ & $-0.02\%$ & $1$ & $0$ \\

$10$ & Zero weights in weighted sparse ID list feature for $100\%$ of examples & 
 $1$ & $ 0.24\%$ & $1$ & $1$ \\

$11$ & $x \to \lfloor x/10\rfloor$ in encoded representation for 2 sparse-encoded embedding features & 
 $2$ & $ -8.81\%$ & $2$ & N/A \\
\bottomrule
\end{tabularx}
\end{table}

Table~\ref{tbl:gfi-mfc-precision-recall} shows our results. We use $K=30$ to measure the recall of the two approaches. We define two types of recall:
\begin{enumerate}
    \item \textbf{Recall}: The overall recall is the percentage of corrupted features that were included in the top $K$ features surfaced to an expert for triaging. The features are ranked based on the shift in importance between control and anomaly data. The importance is measured using the GFI and MFC for the two approaches, respectively.
    \item \textbf{At-Least-One Recall}: At-least-one recall is the percentage of cases where at least one of the corrupted features were included in the top 30. This metric is motivated by our experience where multiple features being corrupted at the same time are typically triaged to a common root cause.
\end{enumerate}

The overall recall is $68\%$ for the XAI approach versus $35\%$ for the MFC approach. The at-least-one recall for the XAI approach is $100\%$ compared to $50\%$ for the MFC approach. The XAI approach is $100\%$ effective in root causing feature value distribution change. In particular, the XAI approach appears to work well
even with significant feature drift including cases 2 and 11, where the model has not been trained on the anomaly data distribution. For coverage drop corruptions, the XAI approach misses features that have \emph{a priori} low GFI (unimportant) when they become corrupted along with important features. When \emph{a priori} unimportant features experience coverage drop by themselves, we expect the average prediction change to be low, and for the calibration to remain close to 1. For features with GFI below the $50$th percentile, we confirmed that a total coverage drop results in average and maximum prediction changes of $0.12\%$ and $1\%$, respectively.

\begin{figure*}[t!]
    \centering
    \begin{subfigure}[t]{0.6\linewidth}
        \centering
        \includegraphics[width=\linewidth]{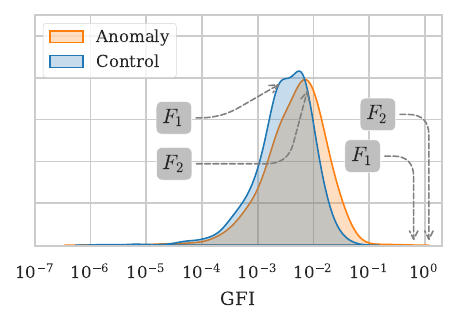}
        \caption{Model without layer normalization}
    \end{subfigure} \\
    \begin{subfigure}[t]{0.6\linewidth}
        \centering
        \includegraphics[width=\linewidth]{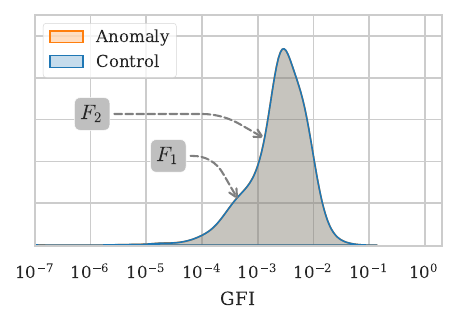}
        \caption{Model with layer normalization}
    \end{subfigure}
    \caption{Comparison of GFI distribution for control and anomaly data for a model with and without layer normalization. $F_1$ and $F_2$ are two sparse-encoded embedding features whose encoded representation is corrupted by dividing each value by $10$.}
    \label{fig:embedding-corruption-gfi}
\end{figure*}

Case 11 is particularly interesting because the 2 sparse-encoded embedding features that experienced corruption have low \emph{a priori} GFI. The two corrupted features, $F_1$ and $F_2$ are computed as $64$-bit floating point vectors of size $1024\times 1$ by an upstream model, quantized to $16$-bit floating point values, and encoded as $64$-bit integers of size $256\times 1$. The corruption in their encoded representations increases the $\ell^2$-norm of the decoded real-valued vectors by $30,000\times$, which causes a prediction change of $-8.81\%$ from the control. The GFI of the features increase from $0.0029$ to $0.6282$ for $F_1$ and $0.0082$ to $1.1794$ for $F_2$. The two corrupted features rank $1$ and $2$ in the top $K$ features identified using the XAI approach. To improve the resilience of the model to such corruptions, we added layer normalization before the interaction layers between dense and embedding features in the model. Figure~\ref{fig:embedding-corruption-gfi} shows the GFI distributions at control and anomaly for the models before and after adding layer normalization.

Given the effectiveness of the XAI approach for \emph{post hoc} root causing of prediction anomalies, we explored its utility for continuous monitoring and proactive anomaly detection. Computing the LFIs for the control and anomaly data is an embarrassingly parallel operation. However, the aggregation step requires LFIs for all examples. We define a sliding window in which a fixed number of examples enter and leave during a step, and compute the GFIs in a continuous manner. This allows us to alert on significant GFI shifts between a lagging control window and a current anomaly window. We found that it is difficult to isolate feature drift from label and concept drifts through such a continuous monitoring system, which results in low precision of alerting for feature corruptions. Nevertheless, this setup is useful for \emph{post hoc} root causing since we can leverage the pre-computed GFIs to reduce triaging time.

\section{Conclusions}
\label{sec:conclusions}

We demonstrated the use of XAI for root-causing performance degradation in continuously trained machine learning models due to feature corruptions. Our approach uses a Feature Ablation algorithm for attributing features relative to a constant baseline. The local feature importances are globally aggregated for unlabeled data sampled from during and prior to a prediction anomaly, and features are ranked by their global importance shift for triaging. We modeled different types of feature corruptions and showed that our approach is effective in identifying corrupted features compared to a model-feature correlation approach.

\begin{credits}
\subsubsection{\ackname} We would like to thank Yijia Liu, Jing (Maggie) Ma, Zewei Jiang, Seunghak (Sean) Lee and other colleagues for the helpful discussions. We acknowledge Ankit Kumar, Subash Sundaresan, Peng Sun and Prabhakar Goyal for their leadership and support of this work.
\end{credits}

\bibliographystyle{splncs04}
\bibliography{references}

\begin{thebibliography}{10}
\providecommand{\url}[1]{\texttt{#1}}
\providecommand{\urlprefix}{URL }
\providecommand{\doi}[1]{https://doi.org/#1}

\bibitem{Budhathoki2021}
Budhathoki, K., Janzing, D., Bloebaum, P., Ng, H.: {Why did the distribution
  change?} In: Proceedings of the 24th International Conference on Artificial
  Intelligence and Statistics. vol.~130, pp. 1666--1674 (2021)

\bibitem{Decker2023}
Decker, T., Gross, R., Koebler, A., Lebacher, M., Schnitzer, R., Weber, S.H.:
  {The Thousand Faces of Explainable AI Along the Machine Learning Life Cycle:
  Industrial Reality and Current State of Research}. In: Artificial
  Intelligence in HCI: 4th International Conference, AI-HCI 2023, Held as Part
  of the 25th HCI International Conference, HCII 2023, Copenhagen, Denmark,
  July 23-28, 2023, Proceedings, Part I. pp. 184--208 (2023)

\bibitem{Joao2014}
Gama, J.a., \v{Z}liobaitundefined, I., Bifet, A., Pechenizkiy, M., Bouchachia,
  A.: A survey on concept drift adaptation. ACM Computing Surveys
  \textbf{46}(4) (2014)

\bibitem{Haug2021}
Haug, J., Z{\"u}rn, S., El-Jiz, P., Kasneci, G.: {On Baselines for Local
  Feature Attributions}  (2021)

\bibitem{He2023}
He, K.: {Find the Root Cause of Model Issues with Actionable Insights}.
  \url{https://www.fiddler.ai/blog/find-the-root-cause-of-model-issues-with-actionable-insights}
  (2023), accessed: 2024-01-27

\bibitem{He2014}
He, X., Pan, J., Jin, O., Xu, T., Liu, B., Xu, T., Shi, Y., Atallah, A.,
  Herbrich, R., Bowers, S., Candela, J.Q.: {Practical Lessons from Predicting
  Clicks on Ads at Facebook}. In: Proceedings of the Eighth International
  Workshop on Data Mining for Online Advertising. pp.~1--9 (2014)

\bibitem{Kenthapadi2022}
Kenthapadi, K., Lakkaraju, H., Natarajan, P., Sameki, M.: {Model Monitoring in
  Practice: Lessons Learned and Open Challenges}. In: Proceedings of the 28th
  ACM SIGKDD Conference on Knowledge Discovery and Data Mining. pp. 4800--4801
  (2022)

\bibitem{Klaise2020}
Klaise, J., Looveren, A.V., Cox, C., Vacanti, G., Coca, A.: Monitoring and
  explainability of models in production (2020)

\bibitem{Koebler2023}
Koebler, A., Decker, T., Lebacher, M., Thon, I., Tresp, V., Buettner, F.:
  {Towards Explanatory Model Monitoring}. In: Workshop: XAI in Action: Past,
  Present, and Future Applications, NeurIPS (2023)

\bibitem{Kokhilyan2020}
Kokhlikyan, N., Miglani, V., Martin, M., Wang, E., Alsallakh, B., Reynolds, J.,
  Melnikov, A., Kliushkina, N., Araya, C., Yan, S., Reblitz-Richardson, O.:
  {Captum: A unified and generic model interpretability library for PyTorch}
  (2020)

\bibitem{Mougan2023}
Mougan, C., Broelemann, K., Kasneci, G., Tiropanis, T., Staab, S.: {Explanation
  Shift}: Detecting distribution shifts on tabular data via the explanation
  space. In: Workshop on Distribution Shifts: Connecting Methods and
  Applications, NeurIPS (2023)

\bibitem{Nigenda2022}
Nigenda, D., Karnin, Z., Zafar, M.B., Ramesha, R., Tan, A., Donini, M.,
  Kenthapadi, K.: {Amazon SageMaker Model Monitor: A System for Real-Time
  Insights into Deployed Machine Learning Models}. In: Proceedings of the 28th
  ACM SIGKDD Conference on Knowledge Discovery and Data Mining. pp. 3671--3681
  (2022)

\bibitem{Sculley2015}
Sculley, D., Holt, G., Golovin, D., Davydov, E., Phillips, T., Ebner, D.,
  Chaudhary, V., Young, M., Crespo, J.F., Dennison, D.: {Hidden Technical Debt
  in Machine Learning Systems}. In: Advances in Neural Information Processing
  Systems 28 (NIPS 2015). vol.~28, pp. 2503--2511 (2015)

\bibitem{Taly2021}
Taly, A., Sato, K.: Monitoring feature attributions: How {Google} saved one of
  the largest {ML} services in trouble.
  \url{https://cloud.google.com/blog/topics/developers-practitioners/monitoring-feature-attributions-how-google-saved-one-largest-ml-services-trouble}
  (2021), accessed: 2024-01-27

\bibitem{Zhang2023}
Zhang, H., Singh, H., Ghassemi, M., Joshi, S.: {``Why did the Model Fail?'':
  Attributing Model Performance Changes to Distribution Shifts}. In:
  Proceedings of the 40th International Conference on Machine Learning (2023)

\bibitem{Zhang2023_1}
Zhang, W., Li, D., Liang, C., Zhou, F., Zhang, Z., Wang, X., Li, R., Zhou, Y.,
  Huang, Y., Liang, D., Wang, K., Wang, Z., Chen, Z., Li, M., Wu, F., Chen, M.,
  Li, H., Wu, Y., Shu, Z., Yuan, M., Reddy, S.: {Scaling User Modeling:
  Large-scale Online User Representations for Ads Personalization in Meta}
  (2023)

\end{thebibliography}

\end{document}